%% file: acl_natbib.tex
\pdfoutput=1

\documentclass[11pt]{article}

\usepackage[]{acl}

\usepackage{times}
\usepackage{latexsym}

\usepackage[T1]{fontenc}

\usepackage[utf8]{inputenc}

\usepackage{microtype}

\usepackage{inconsolata}
\usepackage{pgfplots}     
\usepackage{subfigure}    
\usepackage{tikz}         
\pgfplotsset{compat=1.18} 
\usepgfplotslibrary{groupplots} 
\usepgfplotslibrary[groupplots]
\usetikzlibrary{pgfplots.groupplots}
\usetikzlibrary[pgfplots.groupplots]

\usepackage{multirow}     
\usepackage{booktabs}     
\usepackage{arydshln}     
\usepackage{makecell}     
\usepackage{array}        

\usepackage{amssymb}      
\usepackage{amsmath}      
\usepackage{amsfonts}     

\definecolor{myblue}{RGB}{24,116,205}
\definecolor{myred}{RGB}{178,34,34}
\definecolor{ugreen}{RGB}{124,205,124}
\definecolor{ublue}{RGB}{39,64,139}
\definecolor{myyellow}{RGB}{227,207,87}
\definecolor{myteal}{cmyk}{0.5,0,0.15,0}

\definecolor{qgreen}{RGB}{181,212,101}
\definecolor{qdgreen}{RGB}{099,121,081}
\definecolor{qblue}{RGB}{142,160,199}
\definecolor{qlblue}{RGB}{170,220,224}
\definecolor{qdblue}{RGB}{043,079,125}
\definecolor{qpink}{RGB}{215,144,193}
\definecolor{qorange}{RGB}{235,143,107}
\definecolor{qyellow}{RGB}{250,215,083}
\title{Revisiting Interpolation Augmentation for Speech-to-Text Generation}

\author{Chen Xu\textsuperscript{1}, Jie Wang\textsuperscript{2}, Xiaoqian Liu\textsuperscript{2}, Qianqian Dong\textsuperscript{3}, Chunliang Zhang\textsuperscript{2,4},\\
{\bf Tong Xiao\textsuperscript{2,4},}  {\bf Jingbo Zhu\textsuperscript{2,4}}, {\bf Dapeng Man\textsuperscript{1}\thanks{\ \ Corresponding author.}} and {\bf Wu Yang\textsuperscript{1}} \\
\textsuperscript{1}College of Computer Science and Technology, Harbin Engineering University, Harbin, China\\
  \textsuperscript{2}School of Computer Science and Engineering, Northeastern University, Shenyang, China\\
  \textsuperscript{3}ByteDance \\
  \textsuperscript{4}NiuTrans Research, Shenyang, China \\
  \texttt{\{chen.xu, mandapeng, yangwu\}@hrbeu.edu.cn} \\
  \texttt{\{wangjienlp, liuxiaoqian0319\}@outlook.com}, \texttt{dongqianqian@bytedance.com} \\ 
  \texttt{\{zhangchunliang, xiaotong, zhujingbo\}@mail.neu.edu.cn}
}

\begin{document}
\maketitle
\begin{abstract}
Speech-to-text (S2T) generation systems frequently face challenges in low-resource scenarios, primarily due to the lack of extensive labeled datasets. 
One emerging solution is constructing virtual training samples by interpolating inputs and labels, which has notably enhanced system generalization in other domains. 
Despite its potential, this technique's application in S2T tasks has remained under-explored. 
In this paper, we delve into the utility of interpolation augmentation, guided by several pivotal questions. 
Our findings reveal that employing an appropriate strategy in interpolation augmentation significantly enhances performance across diverse tasks, architectures, and data scales, offering a promising avenue for more robust S2T systems in resource-constrained settings.\footnote{The source code is available at \href{https://github.com/xuchennlp/S2T}{https://github.com/xuchen\\nlp/S2T}.}
\end{abstract}

\section{Introduction}

Recently, neural network-based end-to-end systems have achieved impressive improvements and become the de facto modeling method for speech-to-text (S2T) generation tasks, such as automatic speech recognition (ASR) \cite{Karita_ISCA2019} and automatic speech translation (AST) \cite{xu2023recent}.
These deep learning models typically comprise millions or even billions of parameters and require vast amounts of training data to achieve state-of-the-art performance \cite{zhang2022bigssl}.
For example, leading ASR models demand thousands of hours of training data \cite{Lu_ISCA2020}.
However, the labeling of such extensive datasets leads to significant costs, and models trained on limited data are prone to overfitting, resulting in suboptimal generalization to unseen samples \cite{ying2019overview}.

To enhance generalization capabilities, data augmentation has become a key strategy \cite{shorten2019survey}. 
Existing approaches in S2T can be broadly classified into two categories: online and offline augmentation. 
Online methods, such as SpecAugment \cite{Park_ISCA2019}, enhance regularization by transforming the input representation during training. 
By introducing random noise into input features, these techniques have become standard in S2T tasks. 
Offline methods, on the other hand, boost data diversity by creating large amounts of pseudo-data through original audio distortion \cite{ko2015audio} or synthesis \cite{rosenberg2019speech}. 
Though effective, these offline techniques are separate from the training process, often requiring additional steps and computational resources. This creates a demand for more efficient solutions.

We resort to interpolation augmentation (IPA), also known as Mixup, a notable method first introduced in image classification \cite{zhang2017mixup}. 
IPA mitigates overfitting by constructing virtual samples through linear interpolation of both input features and labels from two randomly selected samples. 
This approach has achieved impressive success across diverse domains, including speech processing \cite{medennikov2018investigation, lam2020mixup, meng2021mixspeech, kang2023learning}, natural language processing \cite{guo2019augmenting, sun2020mixup, xie2023global}, and computer vision \cite{verma2019manifold, wang2023pitfall}.

In the specialized field of speech processing, preliminary studies have explored IPA in speech separation \cite{lam2020mixup, alex2023data} and classification tasks \cite{snyder2017deep, liu2023attention}. However, its application in S2T tasks remains limited and largely unexplored \cite{medennikov2018investigation, meng2021mixspeech, Cheng_Corr2022, zhou2023cmot}. 
The existing work has not yet established clear guidelines on when and how IPA can be optimally leveraged in S2T tasks, leaving a substantial gap in our understanding and application of this promising technique.

In this paper, we examine this question more closely, conducting a series of experiments to answer the following questions:
\begin{enumerate}
    \item[Q1]
    What is the \textbf{appropriate interpolation strategy}, and what distinctions arise between interpolating speech features and text embeddings? (\S\ref{sec:choice})
    \item[Q2] How can IPA create an effective \textbf{combination with existing augmentation techniques}, such as the well-established method SpecAugment? (\S\ref{sec:synergistic})
    \item[Q3] Are there \textbf{specific issues} in applying IPA to S2T tasks, and how can they be addressed? (\S\ref{sec:specific})
    \item[Q4] How does IPA perform \textbf{across various scenarios}? (\S\ref{sec:architectures})
\end{enumerate}

By probing these questions, we develop an effective IPA method that achieves consistent improvements across two S2T tasks (including ASR and AST), various architectures (including encoder-decoder and encoder-CTC), and diverse data scales (ranging from LibriSpeech 10h to 960h).

\section{Experimental Settings}

Data augmentation methods typically demonstrate greater potential in low-resource scenarios. 
In light of this, we conduct analyses using the LibriSpeech 100h ASR dataset and subsequently apply our findings to various scenarios.
We report results mainly on the test-clean and test-other sets.
The average word error rate (WER) is calculated on the concatenation of all four subsets.

Various existing data augmentation techniques, such as SpecAugment and speed perturbation, have achieved excellent results. 
SpecAugment \cite{Park_ISCA2019}, the most widely employed method in S2T tasks, introduces random noise to the input features through time warping, frequency masking, and time masking. 
Speed perturbation \cite{ko2015audio}, on the other hand, commonly expands the dataset by generating three variations of raw audio with speed factors of 0.9, 1.0, and 1.1, facilitating its integration. 
In our work, the goal of IPA is to not only lead to isolate improvements but to also work orthogonally with these methods. 
Therefore, we first examine scenarios without other augmentations and then explore the effects of their combination.

In the field of S2T, common architectures encompass both encoder-decoder (\textbf{Enc-Dec}) and encoder-CTC (\textbf{Enc-CTC}) designs.
The Enc-Dec model consists of an encoder with 12 Conformer layers and a decoder with 6 Transformer layers, each containing 256 hidden units, 4 attention heads, and 2048 feed-forward sizes. 
Connectionist Temporal Classification (CTC, \citealp{Graves_ACL2006}) multi-task learning is applied on top of the encoder, introducing an additional loss with a weight of 0.3. 
The Enc-CTC model can be viewed as a variant of the Enc-Dec model, containing only an 18-layer Conformer encoder for comparable parameters of about 30M.
It predicts the text purely through CTC, where the weight of the CTC loss is 1.
We initially investigate the effects of IPA on the Enc-Dec model before extending the method to the Enc-CTC model.
More details about the datasets and model settings are described in Appendix \ref{app_A}.

\section{Q1: Choice of Interpolation Strategy}
\label{sec:choice}

In this section, we begin with an overview of the basic implementation of IPA.
Subsequently, we investigate the appropriate interpolation strategy tailored specifically for the field of S2T generation.

\subsection{Definition of IPA}

IPA, commonly known as Mixup \cite{zhang2017mixup}, constructs virtual samples in a vicinal distribution by linearly interpolating both the inputs and labels of two randomly selected samples, thereby enhancing the model's generalization capability.
Considering two samples $(x_i, y_i)$ and $(x_j, y_j)$, where $x$ denotes the input features and $y$ represents the corresponding label.
IPA assembles the new sample as follows:
\begin{eqnarray}
    &x_m = \lambda \cdot x_i + (1 - \lambda) \cdot x_j  \label{eq1} \\
    &y_m = \lambda \cdot y_i + (1 - \lambda) \cdot y_j
\end{eqnarray}
where $\lambda \in [0, 1]$ is a weighting factor drawn from a Beta distribution $\lambda \sim \textnormal{Beta}(\alpha, \alpha)$.

A value of $\alpha$ approaching 0 implies that the generated samples closely resemble either $(x_i, y_i)$ or $(x_j, y_j)$, while a value of $\alpha$ approaching $+\infty$ leads to a more balanced interpolation between the two.
In practical applications, IPA randomly replaces a subset of samples with the interpolated versions in each mini-batch, while leaving the remaining samples untouched. 
The selection ratio $\gamma$ is typically set to 1, indicating the model is trained completely on the interpolated samples.
Both $\alpha$ and $\gamma$ serve as essential hyper-parameters, and finding their optimal values often requires careful empirical exploration.

\subsection{IPA Strategy in S2T}

Building upon the aforementioned framework, we extend our investigation to the application of IPA within the domain of S2T generation, focusing specifically on ASR and AST tasks.

Let a training sample be denoted as $(s, x, y)$, where $s$ denotes the speech features, $x$ denotes the transcription of $s$, and $y$ denotes the translation in the target language in AST, or the transcription in the case of ASR.
When employing an Enc-Dec model, the training objectives encompass the utilization of joint CTC loss to model $x$ at the encoder level, coupled with cross-entropy (CE) loss to model $y$ within the decoder. 
Thus, it can be formulated as:
\begin{eqnarray}
    \mathcal{L_{\textnormal{CTC}}}(h, x) \!\!\!&=&\!\!\! - \log \textnormal{P}_{\textnormal{CTC}} (x | h; \theta_{Enc}) \\
    \mathcal{L_{\textnormal{CE}}}(h, z, y) \!\!\!&=&\!\!\! - \log \textnormal{P}_{\textnormal{CE}} (y | h, z; \theta)
\end{eqnarray}
where $h$ is the output of the encoder, and $z$ is the input embedding of the decoder. $\theta_{Enc}$ and $\theta$ are the model parameters of the encoder and the whole network.
Two hyper-parameters $w_{\textnormal{CTC}}$ and $w_{\textnormal{CE}}$ are introduced to balance CTC and CE loss components:
\begin{eqnarray}
    \mathcal{L} = w_{\textnormal{CTC}} \cdot \mathcal{L_{\textnormal{CTC}}} + w_{\textnormal{CE}} \cdot \mathcal{L_{\textnormal{CE}}}
\end{eqnarray}

To apply the IPA in S2T tasks, several significant distinctions must be noted when compared to conventional classification tasks:
\begin{itemize}
    \item In typical classification models, the architecture usually comprises only the encoder, whereas in the S2T model based on the Enc-Dec architecture, the decoder processes the embedding sequence as input. 
    The feasibility of directly interpolating word embeddings remains an open question.
    \item The label in classification tasks often takes the form of a one-hot category, thereby simplifying the interpolation process, while the S2T tasks present a more complex scenario. 
    Specifically, the training objectives for CTC and CE are discrete text sequences, and the method to interpolate and learn the label effectively remains an open question.
\end{itemize}

To address these challenges, we first design the interpolation strategy grounded in previous studies, followed by an exploration of specific issues.
Consider two arbitrary samples in a batch, denoted as $(s_i, x_i, y_i)$ and $(s_j, x_j, y_j)$. We interpolate the input according to Eq. (\ref{eq1}):
\begin{eqnarray}
    s_m = \lambda \cdot s_i + (1 - \lambda) \cdot s_j
\end{eqnarray}
where we pad the shorter features with zeros to achieve the same length for interpolation.
After obtaining the representation $h_m$ outputted by the encoder,  we calculate the CTC loss with respect to both labels and interpolate them as follows:
\begin{eqnarray}
    \mathcal{L_{\textnormal{CTC}}}(h_m, x_i, x_j) 
    = \lambda \cdot \mathcal{L_{\textnormal{CTC}}}(h_m, x_i) \nonumber \\
    + (1 - \lambda) \cdot \mathcal{L_{\textnormal{CTC}}}(h_m, x_j)
\end{eqnarray}

Employing the widely proven interpolation strategy \citet{zhang2017mixup} in the encoder is natural due to similar designs.
Thereby we focus on the interpolation strategy within the decoder.
A straightforward implementation is similar to the operation in the encoder, which involves interpolating the embeddings $z_i$ and $z_j$ in the input layer of the decoder:
\begin{eqnarray}
    z_m = \lambda \cdot z_i +  (1 - \lambda) \cdot z_j
\end{eqnarray}

Next, we calculate losses with two labels $y_i$ and $y_j$ for interpolation.
The whole procedure is formalized as:
\begin{eqnarray}
    \mathcal{L_{\textnormal{CE}}}(z_m, h_m, y_i, y_j) 
    = \lambda \cdot \mathcal{L_{\textnormal{CE}}} (h_m, z_m, y_i) \nonumber \\
    + (1 - \lambda) \cdot \mathcal{L_{\textnormal{CE}}} (h_m, z_m, y_j)
\end{eqnarray}
For simplicity, we refer to this strategy as embedding interpolation (\textit{EIP}).

However, the preceding approach may lead to a disparity between training and decoding. 
During training, the decoder takes the interpolated embedding sequence as input, whereas it receives only a single embedding sequence during inference. 
To bridge this gap, we investigate an alternative strategy that solely interpolates the encoder input while preserving the original input in the decoder \cite{meng2021mixspeech}.
The loss in this context is calculated as follows:
\begin{eqnarray}
    \mathcal{L_{\textnormal{CE}}}(h_m, z_i, z_j, y_i, y_j)
    = \lambda \cdot \mathcal{L_{\textnormal{CE}}} (h_m, z_i, y_i) \nonumber \\
    + (1 - \lambda) \cdot \mathcal{L_{\textnormal{CE}}}  (h_m, z_j, y_j) 
\end{eqnarray}

In summary, two interpolation strategies have distinct characteristics. 
The first approach leverages simple interpolation operations akin to those in the encoder and contributes to the regularization of the decoder. 
Conversely, the second approach focuses on consistent modeling, bypassing interpolation in the decoder, and may facilitate more stable learning.

\begin{table} [tbp]
\centering
\footnotesize
\begin{tabular}{lcccrrr}
\toprule
Method & $\alpha$ & $\gamma$ & \textit{EIP} & clean & other & Avg. \\
\midrule
Baseline & - & - & - & 11.85 & 30.78 & 20.81 \\
\specialrule{0em}{1pt}{1pt}
\cdashline{1-7}
\specialrule{0em}{1pt}{1pt}
\multirow{6}{*}{IPA}
& 0.2 & 1.0 &  & 10.31 & 25.12 & 17.37 \\
& 2.0 & 0.3 &  & 10.31 & 26.44 & 18.00 \\
& 2.0 & 1.0 &  & 10.14 & \textbf{22.45} & \textbf{15.99} \\
\specialrule{0em}{1pt}{1pt}
\cdashline{2-7}
\specialrule{0em}{1pt}{1pt}
& 0.2 & 1.0 & \multirow{3}{*}{$\surd$} & 10.29 & 25.53 & 17.48 \\
& 2.0 & 0.3 & & 10.40 & 26.67 & 18.02 \\ 
& 2.0 & 1.0 & & \textbf{9.91} & 22.90 & 16.35 \\
\bottomrule
\end{tabular}
\caption{
WER of \textbf{IPA} method applied to the \textbf{Enc-Dec} model \textbf{without} SpecAugment on the LibriSpeech 100h dataset.
}
\label{res_wospec}
\end{table}

We construct experiments to validate these two strategies. 
For the hyper-parameters, we select $\alpha$ from the set $\{0.2, 0.5, 1.0, 2.0\}$ and $\gamma$ from $\{0.3, 1.0\}$.
The partial results presented in Table \ref{res_wospec} illustrate that enhancing noise by increasing either $\alpha$ or $\gamma$ serves to reinforce generalization, leading to significant improvements, particularly on the more noisy test-other set.
In addition, the \textit{EIP} strategy exhibits slightly inferior performance.
This observation aligns with our initial conjecture.

\section{Q2: Combination of Augmentation Techniques }
\label{sec:synergistic}

Although the straightforward application of the IPA method has yielded noticeable improvements, our exploration seeks to combine it with existing data augmentation techniques.

\subsection{Preliminary Results}

Table \ref{res_wspec} presents the results obtained when using SpecAugment.
Compared with the IPA method, SpecAugment is more effective in enhancing performance.
However, excessive interpolation intensity inversely affects the results, leading to performance degradation.
Reducing the values of $\alpha$ and $\gamma$ alleviates this issue, though it yields only modest gains.

Another noteworthy observation is that the \textit{EIP} strategy promotes a more stable training process, despite a decline in performance. 
This phenomenon might be attributed to the inherent sensitivity of the original model to noise, coupled with an apparent deficiency in handling complex input within the decoder.
The enhanced robustness introduced by SpecAugment appears to mitigate this issue, empowering the decoder to handle interpolated input and extract information from the noisy encoder output.

\begin{table} [tbp]
\centering
\footnotesize
\begin{tabular}{lcccrrr}
\toprule
Method & $\alpha$ & $\gamma$ & \textit{EIP} & clean & other & Avg. \\
\midrule
Baseline & - & - & - & 8.51 & 19.05 & 13.63 \\
\specialrule{0em}{1pt}{1pt}
\cdashline{1-7}
\specialrule{0em}{1pt}{1pt}
\multirow{8}{*}{IPA}
& 0.2 & 0.3 &  & 8.45 & \textbf{18.68} & \textbf{13.46}  \\
& 0.2 & 1.0 &  & 8.75 & 19.51 & 13.89  \\
& 2.0 & 0.3 &  & 9.19 & 19.88 & 14.27  \\
& 2.0 & 1.0 &  & 11.01 & 23.48 & 16.88 \\
\specialrule{0em}{1pt}{1pt}
\cdashline{2-7}
\specialrule{0em}{1pt}{1pt}
& 0.2 & 0.3 & \multirow{4}{*}{$\surd$} & \textbf{8.29} & 18.97 & 13.53 \\
& 0.2 & 1.0 & & 8.73 & 19.39 & 13.80 \\
& 2.0 & 0.3 & &  8.71 & 19.01 & 13.65  \\
& 2.0 & 1.0 & & 10.36 & 20.24 & 15.07 \\
\bottomrule
\end{tabular}
\caption{
WER of \textbf{IPA} method applied to the \textbf{Enc-Dec} model \textbf{with} SpecAugment on the LibriSpeech 100h dataset.
}
\label{res_wspec}
\end{table}

\subsection{Why Does the Combination Fail?}

To optimize the combination between SpecAugment and IPA, it is crucial to shed light on the influence of SpecAugment on the IPA approach.
Both two methods function by introducing regularization into the encoder input, targeting a balance to improve the model's ability to generalize without causing it to under-fit.
With the right amount of noise, the model may take longer to reach its best performance but eventually perform better.

However, too much noise may result in troubles, leading to training failures or poor results. 
We think that the noise added by SpecAugment might mess up the interpolation, synthesizing samples that stray too far from the desired vicinal distribution. 
As the original samples are replaced with the interpolated version, it leads to poor learning of the actual data distribution.

\begin{figure} [t!]
\centerline{\includegraphics [width=1.19\columnwidth]{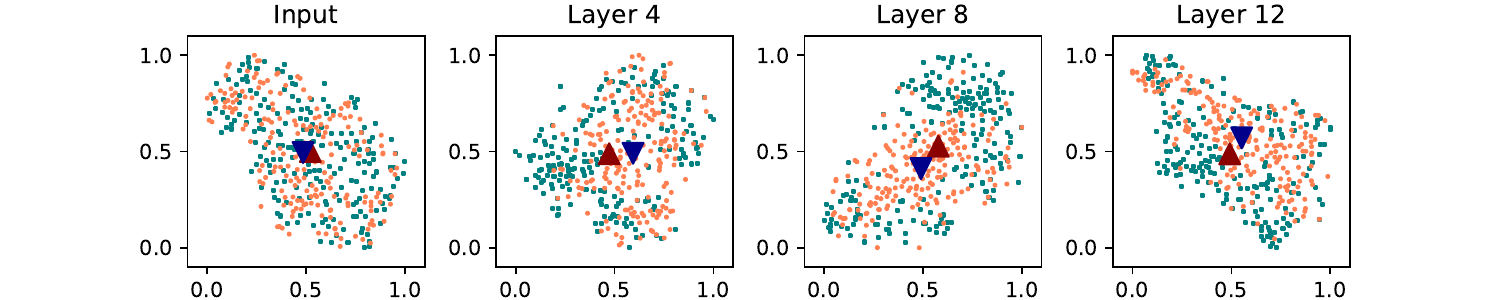}}
\vspace{0.3cm}
\centerline{\includegraphics [width=1.19\columnwidth]{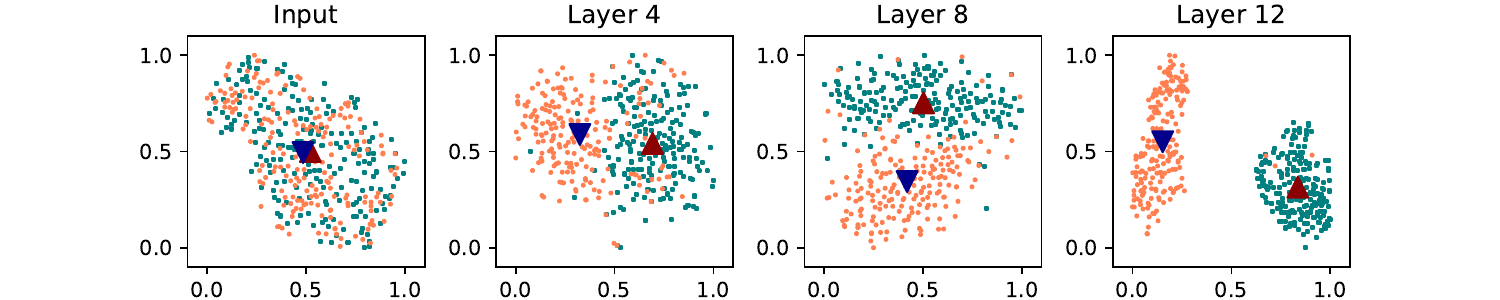}}
\vspace{0.3cm}
\centerline{\includegraphics [width=1.19\columnwidth]{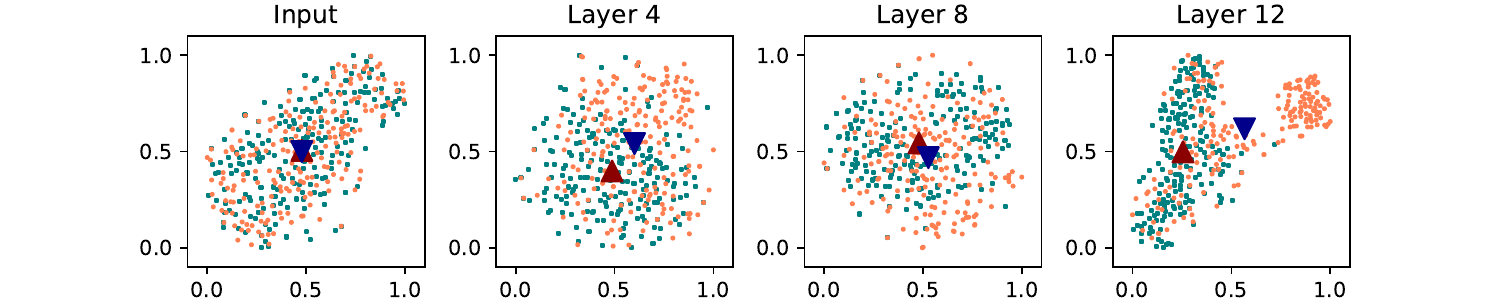}}
\caption{
Visualization of encoder representations of both original (depicted as green squares) and interpolated (depicted as pink circles) samples in the \textbf{IPA} method.
The upper triangle and lower triangle represent the centers of two data distributions, respectively.
The experiment is conducted using the LibriSpeech 100h dataset with an interpolation ratio of $\gamma = 0.3$.
Top: without SpecAugment and $\alpha=2.0$. 
Middle: with SpecAugment and $\alpha=2.0$. 
Bottom: with SpecAugment and $\alpha=0.2$. 
}
\label{visual_ipa}
\end{figure}

To validate our conjecture, we visualize the data distribution of original and interpolated samples by t-SNE.
Figure \ref{visual_ipa} (Top) shows that interpolated samples maintain a similar distribution to that of the original samples when SpecAugment is not employed, even with a large interpolation weight.
However, this distribution uniformity is disrupted with the introduction of SpecAugment, giving rise to an evident discrepancy in distribution, as illustrated in Figure \ref{visual_ipa} (Middle).

Although the initial input representations (the first column) appear similar thanks to the cepstral mean and variance normalization operation, the excessive perturbation caused by SpecAugment leads to a deviation of the interpolated samples from the original empirical distribution during encoding. 
This phenomenon, referred to as \textit{distribution shift}, can be slightly mitigated by diminishing the intensity of the interpolation, thus narrowing the divergence between the two data distributions, as shown in Figure \ref{visual_ipa} (Bottom).
However, traces of the distribution shift persist in the representation at the top layers. 
This inconsistency with the middle layers stems from the influence of the decoder, which we discuss subsequently.

\subsection{Appending-based IPA}

To mitigate the problem of distribution shift identified previously, the key is to prevent the interpolated samples from disturbing the stable learning of the original data distribution.
The "\textit{replace}" operation within the conventional IPA method is revealed to be suboptimal, constraining the magnitude of permissible regularization techniques. 
As an alternative, we introduce an "\textit{appending}" operation into the IPA methodology, referred to as \textbf{AIPA}.
Specifically, for an original batch comprising $n$ samples, AIPA synthesizes $\lceil n \times \gamma \rceil$ interpolated samples. 
These are concatenated with the original batch, resulting in a new batch size of $ \lceil n \times (1 + \gamma) \rceil$ for training.
This simple approach preserves all original samples and generates interpolated ones, thereby safeguarding stable training while simultaneously enabling robust regularization.

\begin{figure} [t!]
\centerline{\includegraphics [width=1.19\columnwidth]{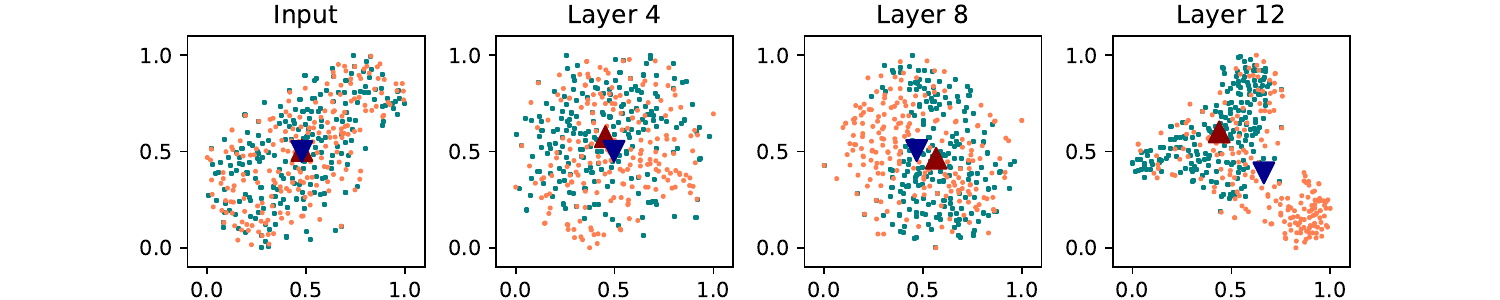}}
\vspace{0.3cm}
\centerline{\includegraphics [width=1.19\columnwidth]{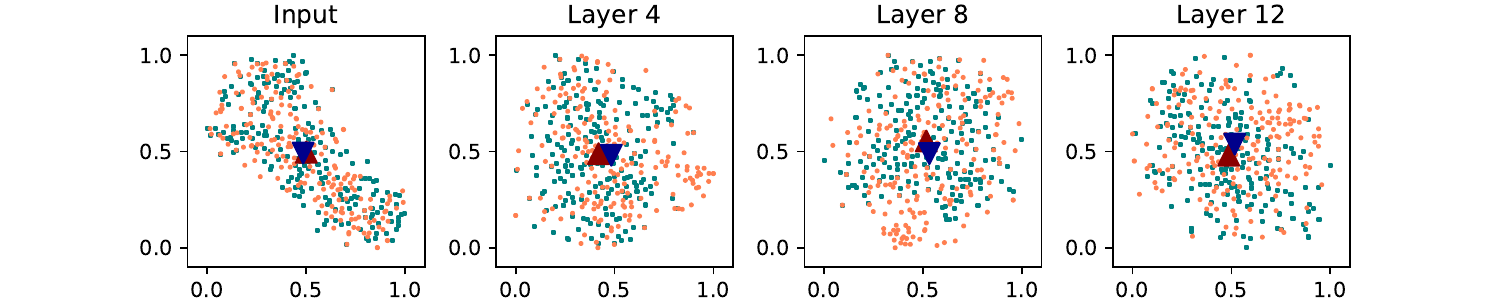}}
\caption{
Similar to Figure \ref{visual_ipa}, visualization of encoder representations in the \textbf{AIPA} method.
Top: Enc-Dec model with SpecAugment, $\alpha=0.2$.  
Bottom: Enc-CTC model with SpecAugment, $\alpha=0.2$. 
}
\label{visual_aipa}
\end{figure}

Moreover, AIPA guarantees exhaustive learning of both the original and vicinal distributions, bridging the divergence between training and inference, as the original samples remain unaltered. 
As depicted in Figure \ref{visual_aipa} (Top), the distance between the two classes of samples has been significantly minimized. 

The experimental results in Table \ref{res_wspec_aipa} further validate these findings.
AIPA yields modest and consistent improvements under the augmentation of varying intensities.
Notably, the \textit{EIP} operation appears to be advantageous. 
This phenomenon can be interpreted as an additional benefit conferred by AIPA, which serves to enhance the robustness of the decoder by introducing controllable regularization. 
Based on these results, we select $\alpha = 0.2$ and $\gamma = 1.0$ as the default hyper-parameters and employ \textit{EIP} operation for the subsequent experiments.

\begin{table} [tbp]
\centering
\footnotesize
\begin{tabular}{lcccrrr}
\toprule
Method & $\alpha$ & $\gamma$ & \textit{EIP} & clean & other & Avg. \\
\midrule
Baseline & - & - & - & 8.51 & 19.05 & 13.63 \\
\specialrule{0em}{1pt}{1pt}
\cdashline{1-7}
\specialrule{0em}{1pt}{1pt}
\multirow{8}{*}{AIPA}
& 0.2 & 0.3 &  & 8.13 & 18.95 & 13.36 \\
& 0.2 & 1.0 &  & 8.01 & 18.52 & 13.05 \\
& 2.0 & 0.3 &  & 8.26 & 18.39 & 13.16 \\
& 2.0 & 1.0 &  & 8.48 & 18.91 & 13.48 \\
\specialrule{0em}{1pt}{1pt}
\cdashline{2-7}
\specialrule{0em}{1pt}{1pt}
& 0.2 & 0.3 & \multirow{4}{*}{$\surd$} & 8.45 & 18.72 & 13.25 \\
& 0.2 & 1.0 & & 7.91 & \textbf{18.14} & \textbf{12.79} \\
& 2.0 & 0.3 & & 8.30 & 18.57 & 13.27  \\
& 2.0 & 1.0 & & \textbf{7.88} & 18.17 & 12.95  \\
\bottomrule
\end{tabular}
\caption{
WER of \textbf{AIPA} method applied to the \textbf{Enc-Dec} model \textbf{with} SpecAugment on the LibriSpeech 100h dataset.
}
\label{res_wspec_aipa}
\end{table}

\input{model}

\section{Q3: Resolution of Specific Issues}
\label{sec:specific}

While the current method does achieve stable effects, the enhancements are relatively modest. 
This section delves into further optimization by addressing the specific issues when employing AIPA in S2T tasks.

In the standard implementation, interpolated samples are given the dual responsibility of predicting two corresponding text sequences in both CTC and CE losses. 
However, this strategy might introduce a risk of ambiguity in the decision boundaries, potentially leading to an over-smoothed model. 
This risk is notably amplified during CTC learning, where the likelihood of a particular transcript $x$ given the hidden representations $h$ is obtained by summing over the probabilities of all feasible alignment paths $\Phi(x)$ between the speech $s$ and $x$:
\begin{eqnarray}
\textrm{P}_{\rm CTC}(x|h) = \sum_{\pi \in \Phi(x)} \textrm{P}(\pi | h)
\end{eqnarray}
This implies that each representation is required to cater to a multiplicity of labels, which substantially complicates the ideal predicted distribution, making it challenging to converge and somewhat counter-intuitive.
Therefore, the design of appropriate training objectives for interpolated samples is pivotal.

We propose constraint objective space (COS), which facilitates CTC learning by replacing the complex traversal with deterministic labels.
Rather than computing the best alignment by the model \cite{xu2023ctc}, we take the predicted distribution of the original samples as the objective of the interpolated samples for efficiency.
Specifically, we calculate the COS loss as follows:
\begin{eqnarray}
    \mathcal{L_{\textnormal{CTC}}^{\textnormal{COS}}} (h_m, h) =
    &\!\!-&\!\!\!\! \sum_{m=1}^{T}\sum_{k=1}^{|V|} \textnormal{P}(\pi_m=v^k|h) \nonumber \\
    &\!\!\times&\!\!\! \log \textnormal{P}(\pi_m=v^k|h_m)
\end{eqnarray}
where $T$ represents the length of $h$, and $V$ denotes the vocabulary.
Drawing a parallel to learning on text labels, we formulate the interpolation of the losses as follows:
\begin{eqnarray}
    \mathcal{L_{\textnormal{CTC}}^{\textnormal{COS}}} (h_m, h_i, h_j)
    = \lambda \cdot \mathcal{L_{\textnormal{CTC}}^{\textnormal{COS}}} (h_m, h_i) \nonumber \\
    + (1 - \lambda) \cdot \mathcal{L_{\textnormal{CTC}}^{\textnormal{COS}}} (h_m, h_j)
\end{eqnarray}

In this framework, the original samples act as a \textit{teacher}, guiding the more accessible learning process of the interpolated \textit{student} \cite{Hinton_Corr2015}. 
This distribution offers detailed information across the entire vocabulary, and importantly, the training objective becomes more deterministic, thereby simplifying the learning process.
The final design of AIPA with COS is depicted in Figure \ref{model}.

Similarly, this strategy can be extended to the cross-entropy (CE) loss, denoted by $\mathcal{L_{\textnormal{CE}}^{\textnormal{COS}}}$.
The final training objective thus takes the form:
\begin{eqnarray}
    \mathcal{L} 
    &=& w_{\textnormal{CTC}} \cdot \mathcal{L_{\textnormal{CTC}}} + w_{\textnormal{CTC}}^{\textnormal{COS}} \cdot \mathcal{L_{\textnormal{CTC}}^{\textnormal{COS}}} \nonumber \\
    &+& w_{\textnormal{CE}} \cdot \mathcal{L_{\textnormal{CE}}} + w_{\textnormal{CE}}^{\textnormal{COS}} \cdot \mathcal{L_{\textnormal{CE}}^{\textnormal{COS}}}
\end{eqnarray}
where $w_{\textnormal{CTC}}^{\textnormal{COS}}$ and $w_{\textnormal{CE}}^{\textnormal{COS}}$ are weights of two COS losses.

\begin{table} [tbp!]
\centering
\footnotesize
\begin{tabular}{lrrrrr}
\toprule
\multirow{2}{*}{Model} & \multicolumn{2}{c}{dev} & \multicolumn{2}{c}{test} & \multirow{2}{*}{Avg.} \\ 
\cmidrule(lr){2-3}\cmidrule(lr){4-5}
\specialrule{0em}{1pt}{1pt}
& clean & other & clean & other &  \\
\midrule
Baseline & 8.20 & 19.13 & 8.51 & 19.05 & 13.63  \\
\specialrule{0em}{1pt}{1pt}
\cdashline{1-6} 
\specialrule{0em}{1pt}{1pt}
AIPA & 7.56 & 17.95 & 7.91 & 18.14 & 12.95          \\
\quad + CTC COS & \textbf{7.11} & \textbf{17.66} & \textbf{7.49} & 17.85 & \textbf{12.43} \\
\quad + CTC COS$^*$ & 7.20 & 17.74 & 7.57 & 17.90 & 12.51  \\
\quad + CE COS  & 7.41 & 17.92 & 7.82 & 17.99 & 12.69 \\
\quad + Both COS & 7.26 & 17.75 & 7.61 & \textbf{17.80} & 12.52 \\
\bottomrule
\end{tabular}
\caption{WER of \textbf{AIPA} method applied to the \textbf{Enc-Dec} model \textbf{with} SpecAugment on the LibriSpeech 100h dataset. The $\alpha$ and $\gamma$ are set to 0.2 and 1, respectively. COS$^*$ indicates using the hard labels.}
\label{res_wspec_aipa_kd}
\end{table}

We present the results in Table \ref{res_wspec_aipa_kd}.
Utilizing COS for CTC training yields an average significant reduction of 1.35 WER points, as this approach simplifies CTC learning by eliminating the need for complex dynamic programming.
Note that the soft training objective is not necessary. The main motivation is to provide simplified labels for stable learning. Replacing the distribution with the one-hot labels by \textit{argmax} operation also achieves obvious effects.
However, the application of COS in CE adversely affects performance. 
We speculate that the CE objective is more straightforward to learn, whereas the COS method might introduce errors.

\section{Q4: Effect under Various Scenarios}
\label{sec:architectures}

We have obtained numerous valuable insights from the ablation studies conducted on the LibriSpeech 100h dataset.
We now extend the application of the aforementioned settings to a broader array of scenarios.

\subsection{Model Architectures}

Combining the above efforts, we develop an effective interpolation augmentation method, which achieves significant improvements in the Enc-Dec architecture. 
The effects are further explored on the Enc-CTC model, with results presented in Table \ref{res_wspec_aipa_ctc}.

Due to the inherent conditional independence assumption of CTC modeling, the baseline model struggles to converge well.
To build more robust configurations, we employ popular techniques to enhance the model's performance. 
Utilizing InterCTC \cite{Lee_ICASSP2021}, additional CTC supervisions are introduced into the intermediate layers, effectively bridging the gap. 
Meanwhile, the prediction-aware encoding (PAE) method \cite{xu2023ctc} integrates self-predicted information, yet only achieves slight improvements due to the limited accuracy of the CTC prediction.

\begin{table} [tbp]
\centering
\resizebox{\columnwidth}{!}{
\footnotesize
\begin{tabular}{lrrrrr}
\toprule
\multirow{2}{*}{Method} & \multicolumn{2}{c}{dev} & \multicolumn{2}{c}{test} & \multirow{2}{*}{Avg.} \\ 
\cmidrule(lr){2-3}\cmidrule(lr){4-5}
\specialrule{0em}{1pt}{1pt}
& clean & other & clean & other &  \\
\midrule
Baseline    & 9.58 & 23.07 & 9.99 & 23.84 & 16.50  \\
\quad + InterCTC & 8.18 & 20.19 & 8.47 & 20.73 & 14.28 \\
\quad\quad + PAE & 8.09 & 19.85 & 8.32 & 20.76 & 14.15 \\
\specialrule{0em}{1pt}{1pt}
\cdashline{1-6} 
\specialrule{0em}{1pt}{1pt}
AIPA                             & 8.77 & 21.41 & 9.07 & 21.75 & 15.14 \\
\quad + CTC COS                   & 7.16 & 17.93 & 7.39 & 18.17 & 12.57 \\
\quad + InterCTC                 & 7.74 & 19.73 & 8.12 & 20.09 & 13.82 \\
\quad\quad + CTC COS              & 7.03 & 17.43 & 7.37 & 17.80 & 12.31 \\
\quad\quad + Both COS             & 6.73 & 17.07 & 6.99 & 17.35 & 11.94 \\
\quad\quad\quad + PAE            & \textbf{6.44} & \textbf{16.49} & \textbf{6.70} & \textbf{16.67} & \textbf{11.49}  \\
\bottomrule
\end{tabular}
}
\caption{WER of \textbf{AIPA} method applied to the \textbf{Enc-CTC} model \textbf{with} SpecAugment on the LibriSpeech 100h dataset. The $\alpha$ and $\gamma$ are set to 0.2 and 1, respectively.}
\label{res_wspec_aipa_ctc}
\end{table}

AIPA yields more substantial improvements on the Enc-CTC model, addressing its inherent fragility. 
The COS method significantly aids CTC learning, resulting in a reduction of 2.57 WER points. 
This result demonstrates the appropriate training objective facilitates convergence effectively.
Within the AIPA method, the intermediate CTC loss is computed similarly to the standard CTC, but its direct use has limited impact. 
However, when coupled with joint COS methods, it achieves gains of 1.88 WER points. 
Finally, thanks to the improved prediction of intermediate CTC, the PAE method also exhibits notable effects.
Combining these methods achieves a remarkable reduction of 2.66 WER points over the baseline model.

Beyond merely improving performance, we also examine the data distribution within the Enc-CTC model, as depicted in Figure \ref{visual_aipa} (Bottom).
Except for applications on various architectures, the settings are consistent with those in the preceding figure.
In the Enc-CTC model, both the original and interpolated samples share the same representation space in the top layers. 
This observation suggests that the distribution shift in the Enc-Dec model is attributable to the behavior of the decoder.
We speculate that the decoder must differentiate between two data distributions to capture information effectively, whereas the CTC objective diminishes this need, thereby maintaining a similar distribution.

\begin{figure}[tbp]
\centering
  \begin{tikzpicture}
    \footnotesize{
      \begin{axis}[
        fill opacity=1,
        fill=orange,
        ymajorgrids,
        xmajorgrids,
        grid style=dashed,
        width=0.45\textwidth,
        height=0.3\textwidth,
        legend columns=4,
        legend entries={
        clean, clean w/ COS, 
        other, other w/ COS },
        legend style={fill opacity=0.5,text opacity =1,
          draw=none,
          line width=1pt,
        },
        legend style={
        at={(0.50,1.25)}, anchor=north,
        nodes={scale=0.8, transform shape}
        },
        xmin=0.2, xmax=2,
        ymin=5, ymax=25,
        xtick={0.2, 0.5, 1.0, 2.0},
        ytick={5, 10, 15, 20,25},
        xlabel=\footnotesize{$\alpha$},
        ylabel=\footnotesize{WER},
        ylabel style={yshift=0.2em},
        scaled ticks=false,
        ]
    \addplot[teal, dashed, line width=1pt] 
    coordinates {
    (0.2,9.07)(0.5,8.92)(1.0,8.78)(2.0,8.90)
    };
    \addplot[teal, line width=1pt] 
    coordinates {
    (0.2,7.39)(0.5,7.50)(1.0,7.74)(2.0,9.27)
    };
    \addplot[orange, dashed, line width=1pt] 
    coordinates {
    (0.2,21.75)(0.5,21.36)(1.0,21.10)(2.0,20.94)
    };
    \addplot[orange, line width=1pt] 
    coordinates {
    (0.2,18.17)(0.5,17.98)(1.0,18.82)(2.0,21.93)
    }; 
    \end{axis}
    }
    \end{tikzpicture}
    \caption{Effects of the hyper-parameters $\alpha$ on Enc-CTC models trained with LibriSpeech 100h dataset.}
    \label{hyper}
\end{figure}

Hyper-parameter $\alpha$ has significant influences on the final performance. 
We illustrate the results of the AIPA method both with and without the COS method in Figure \ref{hyper}. 
AIPA achieves stable results by preserving the original data distribution, and variations in $\alpha$ have only a minor impact. 
However, increasing $\alpha$ negatively affects the efficacy of the COS method. 
A possible explanation is that a larger $\alpha$ results in a more balanced sample interpolation between two original samples, leading to increased COS loss and poor convergence.

In summary, our findings indicate that the IPA technique is particularly well-suited for the Enc-CTC architecture. 
This suitability may stem from multiple factors, such as the baseline model's inherent fragility, the compatibility of continuous features with interpolation, and the elimination of the decoder's influence. 
We will explore these reasons further in future research.

\subsection{Data Scales}

\begin{table}
    \centering
    \resizebox{\columnwidth}{!}{
    \footnotesize
    \begin{tabular}{llccccc}
    \toprule
    \multirow{2}{*}{Dataset} & \multirow{2}{*}{Method} & \multicolumn{2}{c}{dev} & \multicolumn{2}{c}{test} & \multirow{2}{*}{Avg.} \\ 
    \cmidrule(lr){3-4}\cmidrule(lr){5-6}
    \specialrule{0em}{1pt}{1pt}
        &  & clean & other & clean & other &  \\
        \midrule
        \multirow{2}{*}{10h} & Baseline & 35.34 & 51.89 & 35.13 & 53.20 & 43.74 \\
         & AIPA & \textbf{28.34} & \textbf{43.89} & \textbf{28.46} & \textbf{44.76} & \textbf{36.22} \\
        \midrule
         \multirow{2}{*}{50h} & Baseline & 13.10 & 28.40 & 13.48 & 29.46 & 21.03 \\
         & AIPA & \textbf{10.54} & \textbf{22.50} & \textbf{10.84}  & \textbf{23.12} & \textbf{16.64} \\
        \midrule
         \multirow{2}{*}{960h} & Baseline & 3.47 & 9.34 & 3.61 & 9.02 & 6.31 \\
         & AIPA & \textbf{2.91} & \textbf{7.61} & \textbf{3.01}  & \textbf{7.51} & \textbf{5.21} \\
         \bottomrule
    \end{tabular}
    }
    \caption{WER of \textbf{AIPA} method applied to the \textbf{Enc-CTC} model \textbf{with} SpecAugment on the LibriSpeech 10h, 50h, and 960h dataset. InterCTC is used for all models and the COS technique is used in AIPA.}
    \label{tab:res_wspec_aipa_ctc_scales}
\end{table}

By integrating our proposed strategies, we achieve more significant improvements, especially on the noisy other test sets. 
Under the extreme low-resource scenarios of 10h and 50h data, our method achieves substantial reductions of about 4 $\sim$ 6 WER points and boosts the convergence speed effectively.
Even under the high-resource scenario of 960h, AIPA still delivers further improvements.
These findings indicate that the optimized IPA settings are not only effective in low-resource environments but also demonstrate their efficacy in high-resource scenarios.

\subsection{Model Backbones}

\begin{table} [tbp]
\centering
\footnotesize
\begin{tabular}{lcc}
\toprule
Method & Transformer & Conformer  \\
\midrule
Baseline & 6.06 & 7.16  \\
\quad + InterCTC & 5.67 & 5.87 \\
\quad\quad + PAE & 5.32 & 5.81  \\
\specialrule{0em}{1pt}{1pt}
\cdashline{1-3}
\specialrule{0em}{1pt}{1pt}
AIPA & 5.58 & 6.14 \\
\quad + CTC COS & 5.12 & 4.55  \\
\quad\quad + InterCTC & 5.15 & 4.53 \\
\quad\quad\quad + InterCTC COS & 5.05 & 4.35 \\
\quad\quad\quad\quad + PAE & \textbf{4.62} & \textbf{4.27} \\
\bottomrule
\end{tabular}
\caption{
WER of \textbf{AIPA} method applied to the \textbf{Enc-CTC} model \textbf{with} SpecAugment on the AiShell-1 dataset.
}
\label{res_aishell}
\end{table}

We explore the effects of our method with different model backbones on the AiShell-1 ASR dataset, incorporating speed perturbation. 
The results, displayed for both Transformer and Conformer models in Table \ref{res_aishell}, reveal some new insights.
Interestingly, the base Conformer model underperforms its Transformer counterpart, potentially due to underfitting associated with larger model parameters. 
Despite incorporating auxiliary techniques, the Conformer model struggles to converge optimally.

Our proposed method effectively addresses this convergence issue. Notably, employing the COS method specifically for CTC learning offers outstanding regularization and significantly enhances the model's convergence. 
This observation underscores the advantages of our interpolation augmentation method over SpecAugment. 
Across both model architectures, our interpolation strategy yields stable and substantial improvements, illustrating its broadly applicable effectiveness.

\subsection{AST Task}

\begin{table} [tbp]
\centering
\footnotesize
\begin{tabular}{lcc}
\toprule
Method & dev & tst-COMMON  \\
\midrule
Baseline & 25.42 & 26.31 \\
\quad + InterCTC & 26.35 & 26.56 \\
\quad\quad + PAE & \textbf{26.62} & \textbf{26.62} \\
\specialrule{0em}{1pt}{1pt}
\cdashline{1-3}
\specialrule{0em}{1pt}{1pt}
AIPA & 25.85 & 26.38 \\
\quad + CTC COS & 26.04 & 26.75 \\
\quad + CE COS & 26.13 & 26.64 \\
\quad + Both COS & 26.79 & 26.88 \\
\quad + InterCTC & 26.48 & 26.68 \\
\quad\quad + All COS & \textbf{26.92} & \textbf{27.50} \\
\quad\quad\quad + PAE & 26.69 & 27.39 \\
\bottomrule
\end{tabular}
\caption{
BLEU of \textbf{AIPA} method applied to the \textbf{Enc-Dec} model \textbf{with} SpecAugment on the MuST-C En-De ST dataset.
}
\label{res_mustc}
\end{table}

The AST task presents unique challenges due to the substantial modeling complexity involved in handling both cross-modality and cross-lingual mapping.
In this demanding context, the fundamental AIPA method delivers only modest improvements, as shown in Table \ref{res_mustc}. However, with the application of our proposed learning objectives for the interpolated samples, we observe more substantial gains. 

A notable distinction is the effectiveness of the COS method for CE loss. 
This likely stems from the increasing task complexity, where the distribution may be more readily learned by the decoder, thereby easing the training process. 
Remarkably, without resorting to intricate designs, our method achieves a BLEU score of 27.50. This performance is highly competitive, approaching current state-of-the-art results where no additional training data are employed.

\section{Conclusion}
In this paper, we develop a comprehensive exploration of the interpolation augmentation (IPA) method's application in S2T generation.
Our findings provide actionable insights for the effective application of IPA in S2T:
(1) Utilizing IPA alone may not surpass the effectiveness of SpecAugment; a careful combination of both lies in mitigating distribution shift and preserving the learning of original data distribution.
(2) Defining an appropriate training objective for interpolated samples is of paramount importance. 
(3) IPA demonstrates particular compatibility with the Enc-CTC model. 
(4) The appropriate IPA strategy significantly enhances performance across diverse scenarios.

\section*{Limitations}

Although our method demonstrates exceptional performance in various scenarios, there are still some underlying challenges that remain in the follow-up of our work. 
We outline key limitations and propose future directions for improvement:

\begin{itemize}
\item Enhancing stability with diverse hyper-parameters: As depicted in Figure \ref{hyper}, a larger value of $\alpha$ leads to the generation of excessively noisy interpolated samples, adversely affecting the WER. This underscores the need for a more robust IPA method and the determination of universally effective hyper-parameters to ensure broader applicability.
\item Adapting to pre-trained models: The S2T field boasts several influential open-source, pre-trained models such as Wav2vec2.0 \cite{baevski2020wav2vec}, HuBERT \cite{hsu2021hubert}, and Whisper \cite{radford2023robust}. Integrating our IPA method with these established models is a promising avenue that requires thorough validation and exploratory research.
\end{itemize}

\section*{Acknowledgements}

The authors would like to thank anonymous reviewers for their insightful comments.
This work was sponsored by research team project supported by Natural Science Foundation of Heilongjiang (grant no.TD2022F001), NSFC-Xinjiang Joint Fund Key Program (grant no.U2003206), NSFC-Regional Joint Fund Key Program (grant no.U20B2048), National Natural Science Foundation of China (grant no.U21B2019), NSFC-Regional Joint Fund Key Program (grant no.U22A2036), and National Natural Science Foundation of China (grant no.62272127).

\bibliography{acl_natbib}

\appendix
\label{sec:appendix}

\section{Experimental Settings}
\label{app_A}

\subsection{Datasets and Pre-processing}

The datasets are from three benchmarks:

\begin{itemize}
\item {\textbf{LibriSpeech}} is a publicly available read English ASR corpus, which consists of 960-hour training data \cite{Panayotov_ICASSP2015}. 
To assess the performance in both low-resource and high-resource environments, we conduct experiments on LibriSpeech 10h, 50h, 100h, and 960h.
We report results on all four subsets, including dev-clean, dev-other, test-clean, and test-other.
The average word error rate (WER) is calculated on the concatenation of all four subsets.

\item {\textbf{AiSHELL-1}} is a publicly available Chinese Mandarin speech corpus, which consists of 170-hour training data \cite{Hu_aishell}.
We report results WER on both the dev and test sets.

\item {\textbf{MuST-C}} is a multilingual speech translation corpus extracted from the TED talks \cite{Gangi_NAACL2019}. 
We test our method on the MuST-C English-German (En-De) speech translation dataset of 400 hours of speech. We select (and tune) the model on the dev set (Dev) and report the results on the tst-COMMON set (Test).
\end{itemize}

For pre-processing, we follow the standard recipes in fairseq toolkit \cite{Ott_NAACL2019}, which eliminates the utterances of more than 3,000 frames or fewer than 5 frames. 
To explore the impact of integrating another augmentation method, we employ speed perturbation in our experiments conducted on the AiShell-1 dataset.
The extraction of 80-channel Mel filter bank features is carried out using a 25ms window and a stride of 10ms.
For segmentation, we employ SentencePiece \cite{kudo2018sentencepiece} segmentation with a size of 10,000 for the LibriSpeech 100h and MuST-C datasets, 256 for the LibriSpeech 960h dataset. 
And the AiSHELL-1 dataset is segmented using 4231 characters.
For the MuST-C AST dataset, we utilize a shared vocabulary for the source and target languages.

\subsection{Model Settings}

We train the ASR model using the Enc-CTC architecture and AST models with the Enc-Dec architecture.
$\alpha$ and $\gamma$ are set to 0.2 and 1, respectively. 
The weight $w_{\textnormal{CTC}}$ and $w_{\textnormal{CE}}$ for the training objective are set to 0.3 and 1.0 in the encoder-decoder model, while 1.0 and 0.0 in the Enc-CTC model.
And the weight $w_{\textnormal{CTC}}^{\textnormal{COS}}$ and $w_{\textnormal{CE}}^{\textnormal{COS}}$ for the COS method are set to the half of them.
SpecAugment \cite{Park_ISCA2019} is always applied for better results.
Note that our pipeline first applies SpecAugment pre-processing, then performs interpolation augmentation (IPA) on the SpecAugmented samples. This order allows IPA to increase diversity on top of the distortions from SpecAugment

All methods are implemented using the fairseq toolkit.
We employ the Adam optimizer and follow the default learning schedule in fairseq. 
We apply dropout with a rate of 0.1 and label smoothing $\epsilon_{ls} = 0.1$ for regularization.
Note that the feed-forward size is set to 1024 on the LibriSpeech 960h dataset for comparison with previous results.

We do not incorporate pre-training and knowledge distillation techniques during the training process.
We train the model 300 epochs on LibriSpeech 100h and 960h for better convergence and 100 epochs for both AiShell-1 ASR and MuST-C AST datasets.
We early stop training when there is no performance improvement on the development set for 20 consecutive checkpoints.
We report WER/CER and case-sensitive SacreBLEU for ASR and AST tasks, respectively.

\subsection{Augmentation Settings}

In our methodology, SpecAugment is always applied first, followed by sample interpolation. This sequence is based on two key considerations:
\begin{itemize}
    \item SpecAugment is a per-sample operation, whereas IPA can be batch-processed. Applying SpecAugment before IPA results in greater efficiency.
    \item Employing IPA after SpecAugment introduces additional perturbations, potentially enhancing regularization effects. In addition, employing IPA after SpecAugment is easier from the perspective of the code implementation.
\end{itemize}

\end{document}

%% file: model.tex
\definecolor{color1}{RGB}{075,102,173}
\definecolor{color2}{RGB}{098,190,166}
\definecolor{color3}{RGB}{253,186,107}
\definecolor{color4}{RGB}{235,096,070}
\begin{figure}[tbp]
\centering
 
  \begin{tikzpicture}
      \tikzstyle{BackgroundStyle} = [minimum width=22em, minimum height=6.5em,inner sep=4pt,rounded corners=2pt,draw]
      \tikzstyle{SubBackStyle} = [minimum width=6em,minimum height=2.5em,inner sep=4pt,rounded corners=2pt,fill=color3!20]
      \tikzstyle{SBackStyle} = [minimum width=6em,minimum height=3.2em,inner sep=4pt,rounded corners=2pt,fill=color1!20]
      \tikzstyle{xStyle} = [minimum width=0.3em,inner sep=2pt,rounded corners=0.5pt,fill=color3!50,draw]
      \tikzstyle{LossStyle} = [minimum width=4em,minimum height=1.5em,inner sep=2pt,rounded corners=1pt,fill=color4!50,draw]
      \tikzstyle{TransStyle} = [minimum width=5em,minimum height=1.5em,inner sep=2pt,rounded corners=2pt,fill=color3!50]
      \tikzstyle{indexStyle} = [circle,minimum size=1em,fill=color3!40,draw=color3,thick]
    \draw[ultra thick,draw=color2,->](-2,-0.6) -- (-2,0.5);
    \draw[ultra thick,draw=color2,->](2,-1.2) -- (2,1);
    \node[TransStyle,minimum width=14em,fill=color2!50,draw=none](enc)at(0,0) {\footnotesize{Encoder}};
    \node[LossStyle](kdloss)at([yshift=4.2em]enc.center){\footnotesize{COS Loss}};
    \node[SBackStyle](siback) at (-2.1,-1.15) {};
    \node[SBackStyle](sjback) at (-2.1,-2.5) {};
    \node[SBackStyle](smixback) at (2.1,-1.75) {};
    \node[SubBackStyle](xiback) at ([xshift=-6em,yshift=1.5em]kdloss.center){};
    \node[SubBackStyle](xjback) at ([xshift=-6em,yshift=-1.5em]kdloss.center){};
    \node[SubBackStyle](xmixback) at ([xshift=6em]kdloss.center){};
    \node[TransStyle](transj)at([yshift=3.5em]kdloss.center) {\footnotesize{Transcript $x_j$}};
    \node[TransStyle](transi)at([yshift=3.8em]transj.center) {\footnotesize{Transcript $x_i$}};
    \node[LossStyle](ctcloss1)at([xshift=-6em,yshift=2em]transj.center) {\footnotesize{CTC Loss}};
    \node[LossStyle](ctcloss2)at([xshift=6em,yshift=2em]transj.center) {\footnotesize{CTC Loss}};
    \node[circle,minimum size=0.6em,draw,thick] (cir) at ([yshift=-4.8em]enc.center) {};
    \draw[thick]([yshift=-0.05em]cir.north) -- ([yshift=0.05em]cir.south);
    \draw[thick]([xshift=0.05em]cir.west) -- ([xshift=-0.05em]cir.east);
    \node[indexStyle,fill=color1!30,draw=color1,thick](siback)at([yshift=-0.2em]siback.north west){};
    \node[](si) at (siback.center){\footnotesize{{$s_i$}}};
    \node[indexStyle,fill=color1!30,draw=color1,thick](sjback)at([yshift=-0.2em]sjback.north west){};    
    \node[](sj) at (sjback.center){\footnotesize{{$s_j$}}};
    \node[indexStyle,fill=color1!30,draw=color1,thick,minimum size=1.2em](smback)at([yshift=-0.2em]smixback.north east){};    
    \node[](sm) at (smback.center){\footnotesize{{$s_m$}}};
    \node[indexStyle](xiback)at([yshift=-0.2em]xiback.north west){};
    \node[](xi) at (xiback.center){\footnotesize{{$\hat{x_i}$}}};
    \node[indexStyle](xjback)at([yshift=-0.2em]xjback.north west){};
    \node[](xj) at (xjback.center){\footnotesize{{$\hat{x_j}$}}};
    \node[indexStyle,minimum size=1.2em](xmback)at([yshift=-0.2em]xmixback.north east){};
    \node[](xm) at ([yshift=0.1em]xmback.center){\footnotesize{{$\hat{x_m}$}}};
    \node[indexStyle](transiback)at([xshift=1.9em]transi.center){};
    \node[](transxi) at (transiback.center){\footnotesize{{$x_i$}}};
    \node[indexStyle](transjback)at([xshift=1.9em]transj.center){};
    \node[](transxj) at (transjback.center){\footnotesize{{$x_j$}}};
    
    \node[](p1) at ([xshift=-1.6em,yshift=2.2em]cir.center){\footnotesize{$\lambda$}};
    \node[](p2) at ([xshift=-1.5em,yshift=-2.4em]cir.center){\footnotesize{$1- \lambda$}};
    \node[](p3) at ([xshift=-1.7em,yshift=2.1em]kdloss.center){\footnotesize{$\lambda$}};
    \node[](p4) at ([xshift=-1.7em,yshift=-2.5em]kdloss.center){\footnotesize{$1- \lambda$}};
    \node[](p5) at ([xshift=4em,yshift=0.5em]transi.center){\footnotesize{$\lambda$}};
    \node[](p6) at ([xshift=4em,yshift=-0.5em]transj.center){\footnotesize{$1- \lambda$}};

\coordinate (sidot1) at (-3,-1.2) {};
\coordinate (sidot1_5) at ([xshift=0.15em,yshift=1.6em]sidot1.center) {};
\coordinate (sidot2) at([xshift=0.3em,yshift=-1.6em]sidot1.center) {};
\coordinate (sidot25) at ([xshift=0.45em,yshift=1.5em]sidot1.center) {};
\coordinate (sidot3) at ([xshift=0.6em,yshift=-1em]sidot1.center) {};
\coordinate (sidot35) at ([xshift=0.75em,yshift=1.3em]sidot1.center) {};
\coordinate (sidot4) at ([xshift=0.9em,yshift=-1.5em]sidot1.center) {};
\coordinate (sidot45) at ([xshift=1.05em,yshift=1.1em]sidot1.center) {};
\coordinate (sidot5) at ([xshift=1.2em,yshift=-1.2em]sidot1.center) {};
\coordinate (sidot55) at ([xshift=1.35em,yshift=0.5em]sidot1.center) {};
\coordinate (sidot6) at ([xshift=1.5em,yshift=-1.2em]sidot1.center) {};
\coordinate (sidot65) at ([xshift=1.65em,yshift=1em]sidot1.center) {};
\coordinate (sidot7) at ([xshift=1.8em,yshift=-0.5em]sidot1.center) {};
\coordinate (sidot75) at ([xshift=1.95em,yshift=0.4em]sidot1.center) {};
\coordinate (sidot8) at ([xshift=2.1em,yshift=-1.4em]sidot1.center) {};
\coordinate (sidot85) at ([xshift=2.25em,yshift=0.7em]sidot1.center) {};
\coordinate (sidot9) at ([xshift=2.4em,yshift=-0.8em]sidot1.center) {};
\coordinate (sidot95) at ([xshift=2.55em,yshift=0.9em]sidot1.center) {};
\coordinate (sidot10) at ([xshift=2.7em,yshift=-0.2em]sidot1.center) {};
\coordinate (sidot105) at ([xshift=2.85em,yshift=1.5em]sidot1.center) {};
\coordinate (sidot11) at ([xshift=3em,yshift=-0.6em]sidot1.center) {};
\coordinate (sidot115) at ([xshift=3.15em,yshift=0.5em]sidot1.center) {};
\coordinate (sidot12) at ([xshift=3.3em,yshift=-1.4em]sidot1.center) {};
\coordinate (sidot125) at ([xshift=3.45em,yshift=1.2em]sidot1.center) {};
\coordinate (sidot13) at ([xshift=3.6em,yshift=-0.8em]sidot1.center) {};
\coordinate (sidot135) at ([xshift=03.75em,yshift=0.7em]sidot1.center) {};
\coordinate (sidot14) at ([xshift=3.9em,yshift=-1em]sidot1.center) {};
\coordinate (sidot145) at ([xshift=4.05em,yshift=0.6em]sidot1.center) {};
\coordinate (sidot15) at ([xshift=4.2em,yshift=-1.5em]sidot1.center) {};
\coordinate (sidot155) at ([xshift=4.35em,yshift=0.5em]sidot1.center) {};
\coordinate (sidot16) at ([xshift=4.5em]sidot1.center) {};
\draw[rounded corners=0.15cm,semithick] (sidot1.center) --(sidot1_5.center) -- (sidot2.center) --(sidot25.center) -- (sidot3.center) -- (sidot35.center) --(sidot4.center) -- (sidot45.center) --(sidot5.center) -- (sidot55.center) --(sidot6.center) -- (sidot65.center) --(sidot7.center) -- (sidot75.center) --(sidot8.center)-- (sidot85.center) --(sidot9.center) -- (sidot95.center) --(sidot10.center) -- (sidot105.center) --(sidot11.center) --(sidot115.center) -- (sidot12.center) -- (sidot125.center) --(sidot13.center) -- (sidot135.center) --(sidot14.center) --(sidot145.center) -- (sidot15.center) --(sidot155.center) -- (sidot16.center);
\draw[->,thick,rounded corners=0.1em]([xshift=0.86em]sidot16.east) -- ([xshift=3.3em]sidot16.east) -- ([yshift=0.1em]cir.north);

\coordinate (sjdot1) at (-3,-2.5) {};
\coordinate (sjdot1_5) at ([xshift=0.15em,yshift=-0.6em]sjdot1.center) {};
\coordinate (sjdot2) at([xshift=0.3em,yshift=0.6em]sjdot1.center) {};
\coordinate (sjdot25) at ([xshift=0.45em,yshift=-0.7em]sjdot1.center) {};
\coordinate (sjdot3) at ([xshift=0.6em,yshift=0.7em]sjdot1.center) {};
\coordinate (sjdot35) at ([xshift=0.75em,yshift=-0.6em]sjdot1.center) {};
\coordinate (sjdot4) at ([xshift=0.9em,yshift=0.6em]sjdot1.center) {};
\coordinate (sjdot45) at ([xshift=1.05em,yshift=-0.8em]sjdot1.center) {};
\coordinate (sjdot5) at ([xshift=1.2em,yshift=0.9em]sjdot1.center) {};
\coordinate (sjdot55) at ([xshift=1.35em,yshift=-1em]sjdot1.center) {};
\coordinate (sjdot6) at ([xshift=1.5em,yshift=1em]sjdot1.center) {};
\coordinate (sjdot65) at ([xshift=1.65em,yshift=-0.4em]sjdot1.center) {};
\coordinate (sjdot7) at ([xshift=1.8em,yshift=0.8em]sjdot1.center) {};
\coordinate (sjdot75) at ([xshift=1.95em,yshift=-0.8em]sjdot1.center) {};
\coordinate (sjdot8) at ([xshift=2.1em,yshift=0.6em]sjdot1.center) {};
\coordinate (sjdot85) at ([xshift=2.25em,yshift=-1.1em]sjdot1.center) {};
\coordinate (sjdot9) at ([xshift=2.4em,yshift=0.8em]sjdot1.center) {};
\coordinate (sjdot95) at ([xshift=2.55em,yshift=-0.5em]sjdot1.center) {};
\coordinate (sjdot10) at ([xshift=2.7em,yshift=1.2em]sjdot1.center) {};
\coordinate (sjdot105) at ([xshift=2.85em,yshift=-1.3em]sjdot1.center) {};
\coordinate (sjdot11) at ([xshift=3em,yshift=0.6em]sjdot1.center) {};
\coordinate (sjdot115) at ([xshift=3.15em,yshift=-0.5em]sjdot1.center) {};
\coordinate (sjdot12) at ([xshift=3.3em,yshift=0.4em]sjdot1.center) {};
\coordinate (sjdot125) at ([xshift=3.45em,yshift=-0.5em]sjdot1.center) {};
\coordinate (sjdot13) at ([xshift=3.6em,yshift=0.8em]sjdot1.center) {};
\coordinate (sjdot135) at ([xshift=03.75em,yshift=-0.3em]sjdot1.center) {};
\coordinate (sjdot14) at ([xshift=3.9em,yshift=0.6em]sjdot1.center) {};
\coordinate (sjdot145) at ([xshift=4.05em,yshift=-0.8em]sjdot1.center) {};
\coordinate (sjdot15) at ([xshift=4.2em,yshift=1.1em]sjdot1.center) {};
\coordinate (sjdot155) at ([xshift=4.35em,yshift=-0.7em]sjdot1.center) {};
\coordinate (sjdot16) at ([xshift=4.5em]sjdot1.center) {};
\draw[rounded corners=0.15cm,semithick] (sjdot1.center) --(sjdot1_5.center) -- (sjdot2.center) --(sjdot25.center) -- (sjdot3.center) -- (sjdot35.center) --(sjdot4.center) -- (sjdot45.center) --(sjdot5.center) -- (sjdot55.center) --(sjdot6.center) -- (sjdot65.center) --(sjdot7.center) -- (sjdot75.center) --(sjdot8.center)-- (sjdot85.center) --(sjdot9.center) -- (sjdot95.center) --(sjdot10.center) -- (sjdot105.center) --(sjdot11.center) --(sjdot115.center) -- (sjdot12.center) -- (sjdot125.center) --(sjdot13.center) -- (sjdot135.center) --(sjdot14.center) --(sjdot145.center) -- (sjdot15.center) --(sjdot155.center) -- (sjdot16.center);
\draw[->,thick,rounded corners=0.1em]([xshift=0.86em]sjdot16.east) -- ([xshift=3.3em]sjdot16.east) -- ([yshift=-0.1em]cir.south);

\coordinate (smixdot1) at (1.3,-1.8) {};
\coordinate (smixdot1_5) at ([xshift=0.15em,yshift=0.5em]smixdot1.center) {};
\coordinate (smixdot2) at([xshift=0.3em,yshift=-0.5em]smixdot1.center) {};
\coordinate (smixdot25) at ([xshift=0.45em,yshift=0.5em]smixdot1.center) {};
\coordinate (smixdot3) at ([xshift=0.6em,yshift=-0.5em]smixdot1.center) {};
\coordinate (smixdot35) at ([xshift=0.75em,yshift=0.6em]smixdot1.center) {};
\coordinate (smixdot4) at ([xshift=0.9em,yshift=-0.6em]smixdot1.center) {};
\coordinate (smixdot45) at ([xshift=1.05em,yshift=0.6em]smixdot1.center) {};
\coordinate (smixdot5) at ([xshift=1.2em,yshift=-0.6em]smixdot1.center) {};
\coordinate (smixdot55) at ([xshift=1.35em,yshift=0.8em]smixdot1.center) {};
\coordinate (smixdot6) at ([xshift=1.5em,yshift=-0.8em]smixdot1.center) {};
\coordinate (smixdot65) at ([xshift=1.65em,yshift=1em]smixdot1.center) {};
\coordinate (smixdot7) at ([xshift=1.8em,yshift=-1em]smixdot1.center) {};
\coordinate (smixdot75) at ([xshift=1.95em,yshift=1.4em]smixdot1.center) {};
\coordinate (smixdot8) at ([xshift=2.1em,yshift=-1.3em]smixdot1.center) {};
\coordinate (smixdot85) at ([xshift=2.25em,yshift=1.5em]smixdot1.center) {};
\coordinate (smixdot9) at ([xshift=2.4em,yshift=-1.4em]smixdot1.center) {};
\coordinate (smixdot95) at ([xshift=2.55em,yshift=1.2em]smixdot1.center) {};
\coordinate (smixdot10) at ([xshift=2.7em,yshift=-1.1em]smixdot1.center) {};
\coordinate (smixdot105) at ([xshift=2.85em,yshift=1em]smixdot1.center) {};
\coordinate (smixdot11) at ([xshift=3em,yshift=-1.1em]smixdot1.center) {};
\coordinate (smixdot115) at ([xshift=3.15em,yshift=0.5em]smixdot1.center) {};
\coordinate (smixdot12) at ([xshift=3.3em,yshift=-1em]smixdot1.center) {};
\coordinate (smixdot125) at ([xshift=3.45em,yshift=1.2em]smixdot1.center) {};
\coordinate (smixdot13) at ([xshift=3.6em,yshift=-1.2em]smixdot1.center) {};
\coordinate (smixdot135) at ([xshift=03.75em,yshift=0.7em]smixdot1.center) {};
\coordinate (smixdot14) at ([xshift=3.9em,yshift=-0.7em]smixdot1.center) {};
\coordinate (smixdot145) at ([xshift=4.05em,yshift=0.4em]smixdot1.center) {};
\coordinate (smixdot15) at ([xshift=4.2em,yshift=-0.8em]smixdot1.center) {};
\coordinate (smixdot155) at ([xshift=4.35em,yshift=0.6em]smixdot1.center) {};
\coordinate (smixdot16) at ([xshift=4.5em]smixdot1.center) {};
\draw[rounded corners=0.15cm,semithick] (smixdot1.center) --(smixdot1_5.center) -- (smixdot2.center) --(smixdot25.center) -- (smixdot3.center) -- (smixdot35.center) --(smixdot4.center) -- (smixdot45.center) --(smixdot5.center) -- (smixdot55.center) --(smixdot6.center) -- (smixdot65.center) --(smixdot7.center) -- (smixdot75.center) --(smixdot8.center)-- (smixdot85.center) --(smixdot9.center) -- (smixdot95.center) --(smixdot10.center) -- (smixdot105.center) --(smixdot11.center) --(smixdot115.center) -- (smixdot12.center) -- (smixdot125.center) --(smixdot13.center) -- (smixdot135.center) --(smixdot14.center) --(smixdot145.center) -- (smixdot15.center) --(smixdot155.center) -- (smixdot16.center);
\draw[->,thick]([xshift=0.02em]cir.east) -- ([xshift=1.9em]cir.east);

\node[xStyle,minimum height=1em] (xi11)at([xshift=-8.4em,yshift=1.5em]kdloss.center){};
\node[xStyle,minimum height=1.4em] (xi12)at([xshift=0.6em,yshift=0.2em]xi11.center){};
\node[xStyle,minimum height=0.8em] (xi13)at([xshift=0.6em,yshift=-0.3em]xi12.center){};
\node[xStyle,minimum height=0.4em] (xi14)at([xshift=0.6em,yshift=-0.2em]xi13.center){};
\node[xStyle,minimum height=0.8em] (xi21)at([xshift=3em,yshift=-0.1em]xi11.center){};
\node[xStyle,minimum height=0.4em] (xi22)at([xshift=0.6em,yshift=-0.2em]xi21.center){};
\node[xStyle,minimum height=1.2em] (xi23)at([xshift=0.6em,yshift=0.4em]xi22.center){};
\node[xStyle,minimum height=0.6em] (xi24)at([xshift=0.6em,yshift=-0.3em]xi23.center){};
\draw[]([xshift=-0.2em,yshift=0.02em]xi11.south west) -- ([xshift=0.2em,yshift=0.02em]xi14.south east);
\draw[]([xshift=-0.2em,yshift=0.02em]xi21.south west) -- ([xshift=0.2em,yshift=0.02em]xi24.south east);
\node[xStyle,minimum height=1em] (xj11)at([xshift=-8.4em,yshift=-1.5em]kdloss.center){};
\node[xStyle,minimum height=0.4em] (xj12)at([xshift=0.6em,yshift=-0.3em]xj11.center){};
\node[xStyle,minimum height=0.4em] (xj13)at([xshift=0.6em,yshift=0em]xj12.center){};
\node[xStyle,minimum height=1.4em] (xj14)at([xshift=0.6em,yshift=0.5em]xj13.center){};
\node[xStyle,minimum height=0.6em] (xj21)at([xshift=3em,yshift=-0.2em]xj11.center){};
\node[xStyle,minimum height=1em] (xj22)at([xshift=0.6em,yshift=0.2em]xj21.center){};
\node[xStyle,minimum height=0.4em] (xj23)at([xshift=0.6em,yshift=-0.3em]xj22.center){};
\node[xStyle,minimum height=0.6em] (xj24)at([xshift=0.6em,yshift=0.1em]xj23.center){};
\draw[]([xshift=-0.2em,yshift=0.02em]xj11.south west) -- ([xshift=0.2em,yshift=0.02em]xj14.south east);
\draw[]([xshift=-0.2em,yshift=0.02em]xj21.south west) -- ([xshift=0.2em,yshift=0.02em]xj24.south east);
\node[xStyle,minimum height=1em] (xmix11)at([xshift=3.6em]kdloss.center){};
\node[xStyle,minimum height=0.8em] (xmix12)at([xshift=0.6em,yshift=-0.1em]xmix11.center){};
\node[xStyle,minimum height=1.2em] (xmix13)at([xshift=0.6em,yshift=0.2em]xmix12.center){};
\node[xStyle,minimum height=0.6em] (xmix14)at([xshift=0.6em,yshift=-0.3em]xmix13.center){};
\node[xStyle,minimum height=0.4em] (xmix21)at([xshift=3em,yshift=-0.3em]xmix11.center){};
\node[xStyle,minimum height=0.6em] (xmix22)at([xshift=0.6em,yshift=0.1em]xmix21.center){};
\node[xStyle,minimum height=1.0em] (xmix23)at([xshift=0.6em,yshift=0.2em]xmix22.center){};
\node[xStyle,minimum height=0.6em] (xmix24)at([xshift=0.6em,yshift=-0.2em]xmix23.center){};
\draw[]([xshift=-0.2em,yshift=0.02em]xmix11.south west) -- ([xshift=0.2em,yshift=0.02em]xmix14.south east);
\draw[]([xshift=-0.2em,yshift=0.02em]xmix21.south west) -- ([xshift=0.2em,yshift=0.02em]xmix24.south east);

\draw[->,thick,rounded corners=0.1em]([xshift=0.4em]xi24.north east) -- ([xshift=2.78em]xi24.north east) -- ([xshift=-0.6em,yshift=0.1em]kdloss.north);
\draw[->,thick,rounded corners=0.1em]([xshift=0.4em]xj24.south east) -- ([xshift=2.78em]xj24.south east) -- ([xshift=-0.6em,yshift=-0.1em]kdloss.south);
\draw[->,thick,rounded corners=0.1em] ([xshift=1em]kdloss.east) -- ([xshift=0.1em]kdloss.east);

\draw[thick,->]([yshift=-2em]ctcloss1.south) -- ([yshift=-0.1em]ctcloss1.south);
\draw[thick,->]([yshift=-3.5em]ctcloss2.south) -- ([yshift=-0.1em]ctcloss2.south);
\draw[->,thick,rounded corners=0.2em]([xshift=-0.6em]xj11.center) -- ([xshift=-1.6em]xj11.center) -- ([xshift=-1.6em,yshift=5em]xj11.center) --([xshift=1.9em,yshift=5em]xj11.center) -- ([xshift=-0.5em,yshift=-0.1em]ctcloss1.south);
\draw[->,thick,rounded corners=0.2em](transj.west) -- ([xshift=-2.96em]transj.west) -- ([xshift=0.5em,yshift=-0.1em]ctcloss1.south);
\draw[->,thick,rounded corners=0.2em](transj.east) -- ([xshift=2.96em]transj.east) -- ([xshift=-0.5em,yshift=-0.1em]ctcloss2.south);
\draw[->,thick,rounded corners=0.2em](transi.west) -- ([xshift=-3em]transi.west) -- ([xshift=0.5em,yshift=0.1em]ctcloss1.north);
\draw[->,thick,rounded corners=0.2em](transi.east) -- ([xshift=3em]transi.east) -- ([xshift=-0.5em,yshift=0.1em]ctcloss2.north);

\end{tikzpicture}
\caption{Encoding process of the AIPA method with COS training.}
\label{model}
\end{figure}